\documentclass[journal]{IEEEtran}
\usepackage{cite}
\usepackage{amssymb}
\usepackage{graphicx}
\usepackage{multicol}
\usepackage{multirow}
\usepackage{subfig}
\usepackage[fleqn]{amsmath}
\usepackage{amsfonts,amssymb}
\graphicspath{{./fig/}}
\usepackage[dvips]{epsfig}
\usepackage{url}
\newcommand{\Tau}{\mathcal{T}}
\usepackage{lipsum}
\usepackage{soul}
\usepackage{xcolor}

\begin{document}

\title{Performance Analysis of Series Elastic Actuator based on Maximum Torque Transmissibility}

\author{Chan Lee, and Sehoon Oh
\thanks{C. Lee, and S. Oh are with the Department of Robotics Engineering, DGIST (Daegu Gyeongbuk Institute of Science and Technology), Daegu, Korea  (e-mail: {\{chanlee, sehoon\}@dgist.ac.kr})}}
\maketitle

\begin{abstract}
The use of the Series Elastic Actuator (SEA) system as an actuator system equipped with a compliant element has contributed not only to advances in human interacting robots but also to a wide range of improvements in the robotics area. Nevertheless, there are still limitations in its performance; the elastic spring that is adopted to provide compliance is considered to limit the actuator performance thus lowering the frequency bandwidth of force/torque generation, and the bandwidth decreases even more when it is supposed to provide large torque.

This weakness is in turn owing to the limitations of motor and motor drives such as torque and velocity limits. In this paper, mathematical tools to analyze the impact of these limitations on the performance of SEA as a transmission system are provided. A novel criterion called Maximum Torque Transmissibility (MTT)is defined to assess the ability of SEA to fully utilize maximum continuous motor torque. Moreover, an original frequency bandwidth concept, maximum torque frequency bandwidth, which can indicate the maximum frequency up to which the SEA can generate the maximum torque, is proposed based on the proposed MTT. The proposed MTT can be utilized as a unique criterion of the performance, and thus various design parameters including the load condition, mechanical design parameters, and controller parameters of a SEA can be evaluated with its use. Experimental results under various conditions verify that MTT can precisely indicate the limitation of the performance of SEA, and that it can be utilized to accurately analyze the limitation of the controller of SEA.

\end{abstract}
 
\section{Introduction}
\label{sec:intro}
The necessity for compliant actuation has emerged as a key technology requirement in the field of robotics in order to achieve safe interactions with humans while achieving given tasks. There are many approaches investigated to achieve safe robot interaction with humans, such as link-mass reduction, impedance control, and increased compliance \cite{Hogan85, Haddadin08, Laffranchi09}.

In the wake of these research approaches, SEA, which embeds a compliant element, has been developed \cite{Pratt95}. Specifically, this element is a spring set in series between the motor and the load, which facilitates force sensing without extra sensors, as it enables control of the force to the load through spring deformation.

SEA has been developed as an ideal force source, which has low output impedance with low reflected inertia and low friction \cite{Robinson99} for safety and high force fidelity. The advantages of SEA over conventional actuators also include energy storage capability, low cost force measurement, low cost transmission, and better force control stability \cite{Pratt02, Sariyildiz16}. Having these beneficial characteristics, SEA has been applied to various robotic devices, such as rehabilitation robots \cite{Vallery08}, humanoid robots \cite{paine15}, quadrupedal robots \cite{hutter12}, robotic prosthesis \cite{Au07, Grimmer11}, and industrial robots \cite{Guizzo12}. As the use of SEA increases, many robotics researchers have demonstrated the performance of SEA as the next generation actuator system, and the results of their studies in this regard have been reported in many recent papers \cite{fujimoto2016advanced, zhu2014design, yu2015human, li2017adaptive, oh2017high, FO2017}.

In spite of the many benefits of SEA, mechanical complexity has been its most problematic issue since its dynamics consist of a combination of motors as well as gears and compliant components. This mechanical complexity in the dynamics of SEA leads to various limitations including the difficulty of high performance controller design \cite{Kong2012, Paine14, wyeth, Vallery07}. A number of research studies have investigated the ability and limit of the mechanical characteristics of SEA to address this problem \cite{Laffranchi13, Kemper10, Calanca15}. 

One of the main limitations of SEA, which is widely accepted among SEA researchers, and has been demonstrated and analyzed by many researchers \cite{Au07, Vallery07, Lagoda10}, is its reduced bandwidth \cite{kooij06} particularly when the magnitude of the force that the SEA is supposed to provide is large \cite{Pratt95, hurst04}.

In addition to this bandwidth limitation, the deterioration of large/maximum torque generation has been an issue in SEA applications. \cite{robinson2000} has analyzed this problem in detail based on the electro-mechanical characteristics of a motor; the torque of a motor decreases as its speed increases, which suggests that the motor velocity limitation can hinder the large/maximum torque generation of SEA. This has been a significant issue in SEA, and several succeeding studies have accepted this idea and utilized it as a design criterion in SEA systems \cite{Au07, Lagoda10, Sensinger06}.

The actual limitation of motor and motor drive, however, does not always follow the constant electro-mechanical characteristics utilized in \cite{robinson2000, Sensinger06}, particularly when it is under current control. Moreover, the analyses in \cite{robinson2000, Sensinger06} did not take into account the controller design nor the load conditions. In other words, the studies did not fully consider the dynamics of SEA, and the results of the studies provided more qualitative discussion than quantitative ones. Therefore, it is difficult to use the results as criterion for the mechanical or control design of SEA. This paper, therefore, proposes a more practical analysis of the limitation of large/maximum torque generation in SEA taking into consideration all the feedback controllers and load conditions.

An accurate and practical analytic tool is proposed to address the large force generation problem of SEA in this paper; a novel criterion, called Maximum Torque Transmissibility (MTT), is proposed to assess the ability of SEA to fully utilize the maximum continuous motor torque. By using the proposed MTT, it can be quantitatively shown that the performance of SEA is either maintained or deteriorated when the desired torque becomes large and how the mechanical design parameters and controller parameters affect large force generation performance.

The proposed MTT is a function of frequency and it can analyze the performance of SEA in the frequency domain, and thus, the accurate frequency bandwidth can be determined. This bandwidth is named maximum torque bandwidth and can specify the frequency bandwidth up to which the transmission of the maximum motor torque is guaranteed.

In the proposed MTT, not only the torque limit but also the velocity limit of the motor drive is taken into account as well, thus providing the comprehensive analysis of the limitation of the SEA performance caused by the motor drive. In addition, the proposed MTT can be applied to force generation during dynamic motions, which has not been studied in previous research.

Focusing on the foregoing, this paper will make the following contributions.
\begin{enumerate}
\item A novel criterion, MTT is proposed to assess the ability of SEA to transmit the maximum motor torque.
\item The maximum torque bandwidth of SEA is defined using the proposed MTT.
\item Using the proposed maximum torque bandwidth as criterion, the influence of the load condition, the mechanical parameters, and the controllers of SEA are evaluated, and design guidelines are given based on it.
\end{enumerate}

This paper is organized as follows. Section \ref{sec:MTT} discusses the necessity and definition of the proposed MTT. The large torque generation problem is first introduced, then the dynamic characteristic of SEA (including environments) is explained. MTT is derived based on the dynamic characteristic of SEA. In Sec. \ref{sec:exp_MTT}, the proposed MTT is verified through experiments. Through the experiments, it is verified that the MTT can identify the relationship between gear ratio and large torque generation capability. Section \ref{sec:MTT_omega} shows that the proposed MTT can be utilized as guide for SEA mechanical and controller design. Concluding discussions are given in Sec. \ref{sec:conclusion}.

\section{Maximum Torque Transmissibility of SEA}\label{sec:MTT}
This section describes the deterioration of force control performance when SEA provides a large force and explains the background that needs to be analyzed to address this phenomenon. To solve this problem, Maximum Torque Transmissibility (MTT), defined as a quantitative criterion that evaluates SEA force control performance considering motor torque limit and velocity limit, is utilized.
\subsection{Large Torque Generation Performance Deterioration}
\begin{figure}[t!]
\begin{center}
\includegraphics[width=0.8\columnwidth]{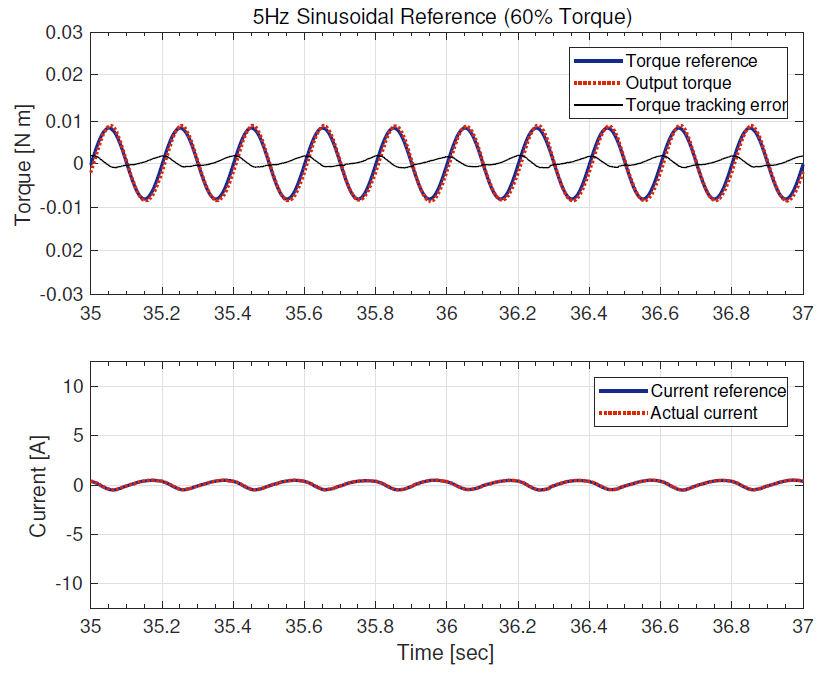}\\
{\footnotesize (a)}\\
\includegraphics[width=0.8\columnwidth]{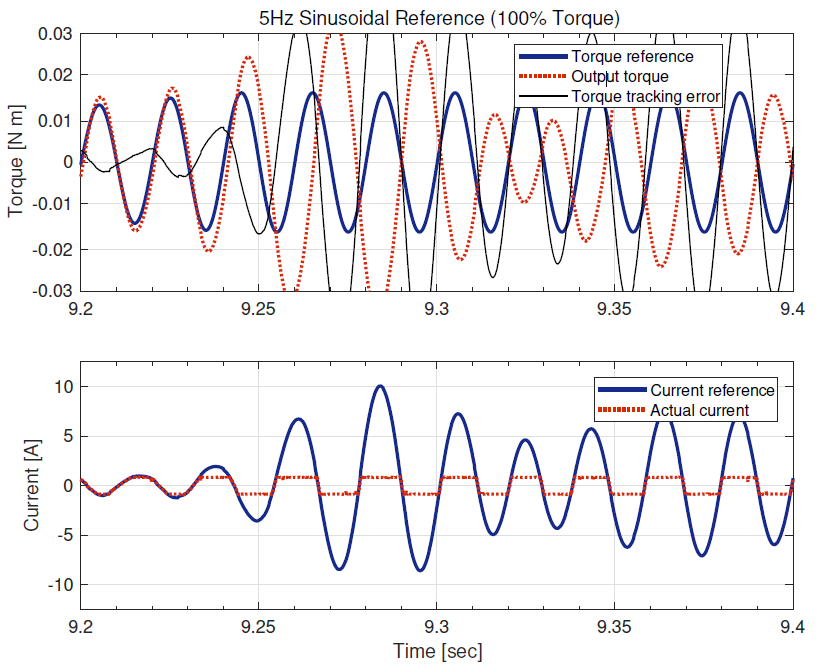}\\
{\footnotesize (b)}
\caption{\footnotesize Force tracking performance with two different desired torque magnitudes. (a) has the desired torque magnitude set to 0.6 $\times$ the gear ratio $\times$ the maximum continuous motor torque, while (b) has the desired torque magnitude set to the gear ratio $\times$ the maximum continuous motor torque. The upper graphs show torque reference tracking results and errors, whereas the lower graphs show the current references and actual currents of the motor driver.}
\label{fig:5Hz_Force}
\end{center}
\end{figure}
As SEA can be considered a transmission system consisting of a spring and a reduction gear, it is supposed to be capable of transmitting the maximum continuous motor torque multiplied by the gear ratio. This torque transmission or generation of SEA is realized through the control of spring deformation, which guarantees the transmission performance up to a certain frequency bandwidth. However, this bandwidth is degenerated when SEA is controlled to provide the maximum torque. 

Figure \ref{fig:5Hz_Force} shows the experimental result of force generation/tracking of SEA, where the force controller is designed to provide up to 8 Hz frequency bandwidth. Five Hz sinusoidal torque patterns with two different magnitudes were applied as the desired torque reference: one was 60$\%$ of the rated motor torque (multiplied by the gear ratio), and the other was 100$\%$ of the rated motor torque. %

The result shows that the SEA failed to generate the desired torque at 5 Hz when the reference magnitude is set to the maximum level. It can be noticed that the controller is designed to guarantee tracking at 5 Hz, which can be verified in Fig. \ref{fig:5Hz_Force}. (a). The measured current that is actually provided to the motor (the bottom figures of Fig. \ref{fig:5Hz_Force}) explains the cause of this performance deterioration; the current is limited by the motor drive when the desired torque magnitude is set large, and fails to track the current reference set by the force controller of the SEA. It then becomes necessary to know when and how this maximum torque generation failure start to happen. 

\subsection{Limitation of Motor and Drive System}\label{sec:motor_limit}
\begin{figure}[t]
\includegraphics[width=0.9\columnwidth]{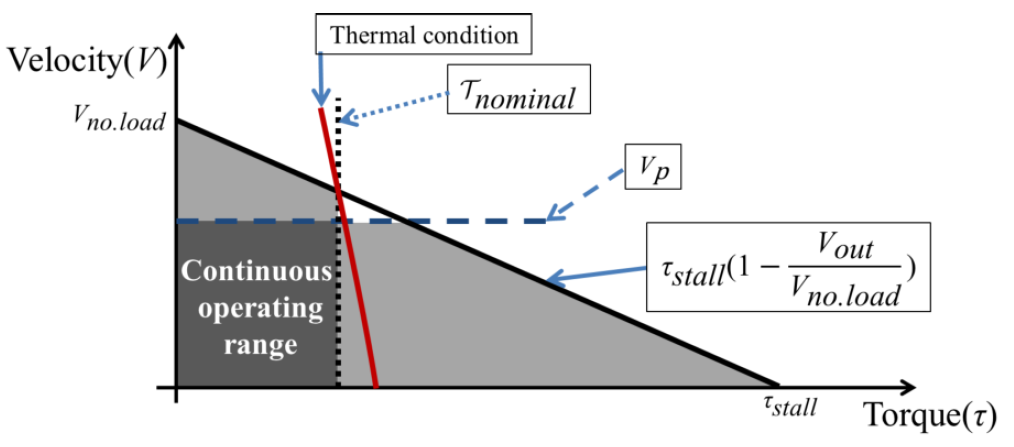}
\caption{Continuous operating range of a DC motor considering various conditions}
\label{fig:op_range}
\end{figure}

In most cases, the limitation of the motor torque and velocity
is determined by the motor drive system, which is far below the torque/speed curve of the motor \cite{motor}. Figure \ref{fig:op_range} shows a typical operation range of a DC motor along with the torque/speed curve of the motor. Notice that the torque limitation (the nominal torque, $\Tau_{nominal}$) and the velocity limitation (the maximum permissible speed $V_p$) are set independently from the torque/speed curve.

In \cite{maxon}, the velocity limit was considered to follow the torque/speed curve (the thick line in Fig. \ref{fig:op_range})implying that it varies with regard to the torque output. The actual torque and velocity limits, however, are not set according to this line but according to constant values independent of each other \cite{motor}. Therefore, the discussion in \cite{maxon} cannot be regarded as effective in practical cases.

The torque and velocity that are generated by a DC or BLDC motor are limited by its electro-mechanical dynamics. In addition to this limitation, the motor driver also sets limitations on the torque and velocity.

The nominal torque of a DC motor is determined by the thermal condition, i.e., the nominal current of the motor is selected so that the winding temperature is kept under the maximum temperature.

In Fig. \ref{fig:op_range}, the thermal condition is depicted as a red solid line on the torque/speed curve, and the nominal current value is determined by the intersection point between the torque/speed curve and the thermal curve. The maximum continuous torque of motor $\Tau_{nominal}$ is determined by the nominal current multiplied by the torque constant.

The motor speed $v_m$ is limited due to various reasons: the mechanical wear and the electro-erosion of brushes and commutators of a brushed DC motor, and the service life of the bearings \cite{maxon}.
Figure \ref{fig:op_range} shows the limitation of the motor velocity which is also known as the maximum permissible speed $V_{p}$. Note that $V_{p}$ is constant and not related to the motor torque output.

These limitations of the motor lead to the limitation in the performance of SEA as a transmission system. It is required that SEA fully utilize the continuous operation range in Fig. \ref{fig:op_range} of the motor as an efficient transmission. The proposed maximum torque transmissibility, which is derived by taking into consideration these motor limitations, can be a criterion to indicate this effectiveness.

\subsection{Generalized Dynamic Model of SEA}\label{sec:dyn_mdl_SEA}
In order to accurately explore and exploit the best performance of SEA, it is required to understand the dynamic characteristics of SEA in terms of motion and force/torque generation.

The mechanical structure of SEA becomes more complicated than a conventional rigid actuator system as SEA consists of several dynamic components such as an electric motor, a spring for elasticity, reduction gears, and the load.

\begin{figure}[b]
\begin{center}
\includegraphics[width=0.9\columnwidth]{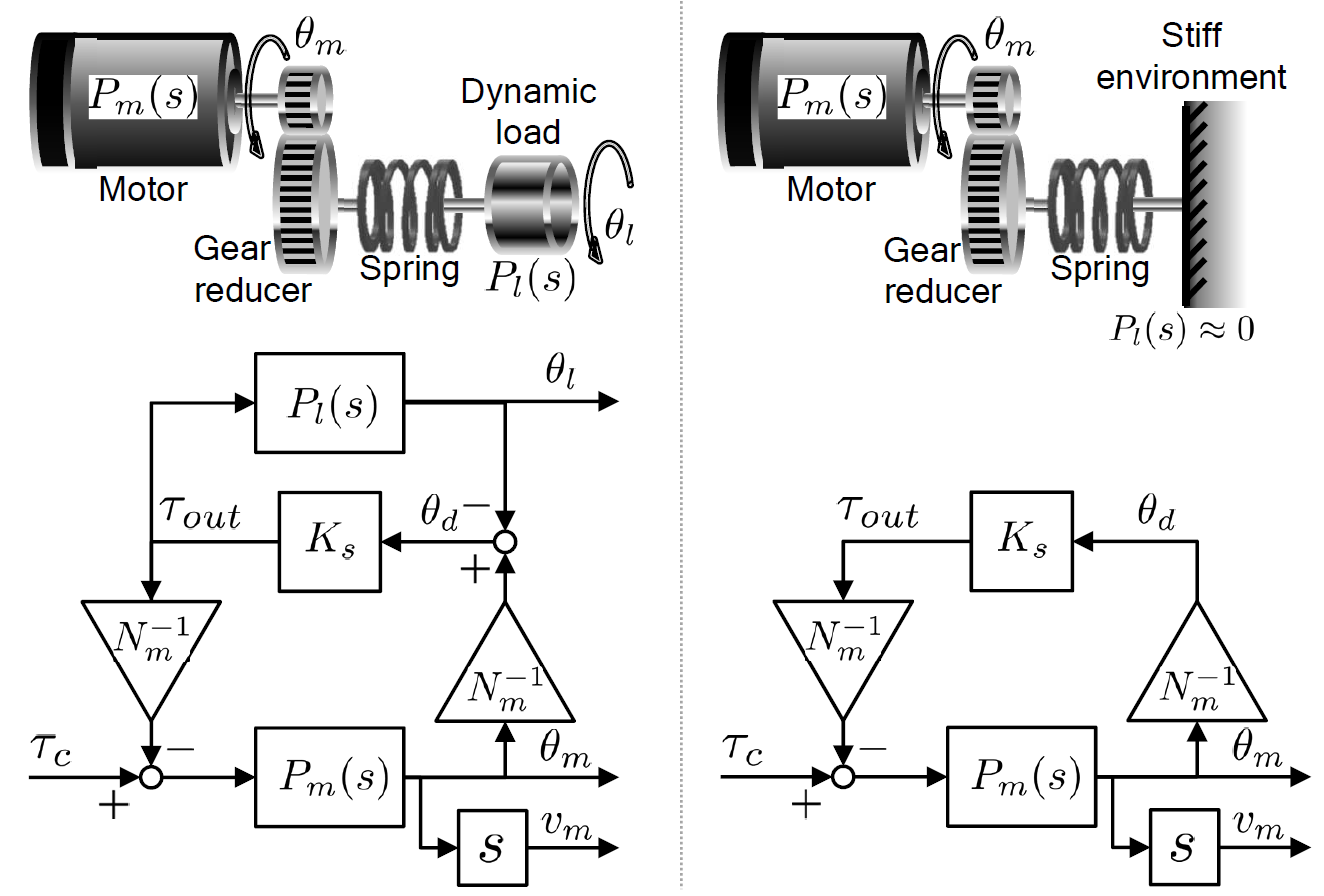}
\caption{\footnotesize Block diagram describing SEA dynamics. The left side figures are cases where SEA moves freely connected to a load, and the right side figures are cases where SEA is in contact with a stiff environment. $P_m(s)$ and $P_l(s)$ represent the dynamics of the motor and the load in the SEA, respectively. $K_s$ is the stiffness of the spring, $N_m$ is the gear ratio, and $\tau_c$ is the motor torque input to the SEA. $\theta_m$ and $\theta_l$ are the angles of the motor and the load, respectively, and their difference $\theta_d = N_m^{-1}\theta_m - \theta_l$ represents the spring deflection. $v_m$ is the motor velocity output.}
\label{fig:SEA_block}
\end{center}
\end{figure}

As the spring connects the motor and the load in SEA, it can be modeled as a two-mass system. The left figures in Fig. \ref{fig:SEA_block} illustrates the dynamic model of SEA as a two-mass system, where the motor $P_m(s)$ and the load $P_l(s)$ are connected through the spring $K_s$. $P_m(s)$ and $P_l(s)$ are modeled as
\begin{equation}
P_{m}(s)=\frac{1}{J_{m}s^2+B_{m}s},P_{l}(s)=\frac{1}{J_{l}s^2+B_{l}s},
\label{eq:motor_load_dyn}
\end{equation}
where $J_{m}$ and $B_{m}$ represent the moment of inertia and the damping coefficient of the motor, respectively, and $J_{l}$ and $B_{l}$ represent those of the load. In this paper, load side dynamics $P_l(s)$ is modeled as (\ref{eq:motor_load_dyn}), whereas SEA can be contacted with a variety of external environments, which can be modeled as closely as possible to a real contact environment \cite{oh2017high}.

The output torque of SEA as a transmission is $\tau_{out}$ which is transmitted to the load, and it is determined by $K_s\theta_d$, the product of the spring deformation and the stiffness, respectively. Therefore, the dynamic characteristic of SEA can be defined as the transfer function from the motor torque $\tau_c$ to the output torque $\tau_{out}$, given as
\begin{equation}
P_{dynamic}(s)=\frac{N_{m}^{-1}K_{s}P_{m}(s)}{1+P_l(s)K_{s}+N_{m}^{-2}P_m(s)K_{s}}.
\label{eq:tf_ol_low}
\end{equation}
This transfer function can be interpreted as the transmissibility of SEA representing the relationship between the input torque and the output torque of SEA \cite{act6030026}.

This derived dynamic characteristic changes when the SEA contacts stiff environments as shown in the right side figures in Fig. \ref{fig:SEA_block}. In this case, SEA is supposed to provide force/torque directly to the environment rather than to generate motions. 
By limiting $J_l$ and $B_l$ in the load dynamics (\ref{eq:tf_ol_low}) to infinity, the transfer function of SEA can be seamlessly shifted to the high impedance environment case, given as follows.
\begin{equation}
P_{static}(s)=\frac{N_{m}^{-1}K_{s}P_{m}(s)}{1+N_{m}^{-2}P_m(s)K_{s}}.
\label{eq:trq_tf_ol}
\end{equation}
Notice that the spring deflection $\theta_{d}$ becomes equal to the motor angle $\theta_{m}$ multiplied by $N_m^{-1}$.
The block diagram of SEA in this case can be derived in the same way, and is given in the right side figures of Fig. \ref{fig:SEA_block}.

The transfer function to the motor velocity $v_m$ is also required when analyzing the maximum torque generation. Based on the model in the left side figures of Fig. \ref{fig:SEA_block}, the transfer function from the motor torque $\tau_c$ to the motor velocity can be derived as
 \begin{equation}
P_V(s)=\frac{P_{m}(s)[1+K_sP_l(s)]s}{1+P_l(s)K_{s}+N_{m}^{-2}P_m(s)K_{s}}.
\label{eq:trq_tf_ol_v}
\end{equation}

In the remainder of the paper, the case when SEA is freely moving with a dynamic load is referred to as the \textit{dynamic load case}, whereas it is called the \textit{static load case} when SEA contacts a stiff environment. In this paper, both cases are considered in the analysis of force generation performance, which is different from other studies where only the static load case was discussed \cite{Au07, Lagoda10, Sensinger06, lee2016maximum}.

\subsection{Control Input and Motor Velocity under Force Feedback Control}\label{sec:controller}
\begin{figure}[b]
\begin{center}
\includegraphics[width=0.8\columnwidth]{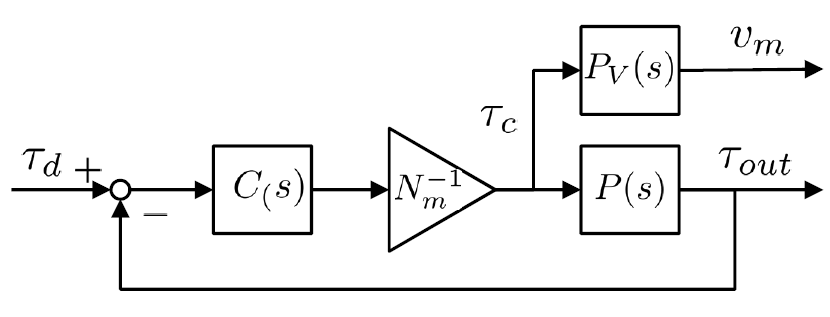}
\caption{\footnotesize Configuration of SEA force feedback control to provide desired torque $\tau_d$ from SEA. $P(s)$ is the transfer function of SEA given in (\ref{eq:tf_ol_low}) or (\ref{eq:trq_tf_ol}) according to the load condition. $P_v(s)$ is the transfer function from control input to the motor output velocity in (\ref{eq:trq_tf_ol_v}). $C(s)$ is a feedback controller.}
\label{fig:SEA_FCFB}
\end{center}
\end{figure}
For SEA to generate the desired torques $\tau_d$, the output torque $\tau_{out}$ (which is spring deformation $\times$ spring stiffness) should be controlled. Figure \ref{fig:SEA_FCFB} illustrates the SEA system $P(s)$ under the force control $C(s)$ to achieve this. Note that $P(s)$ can be modeled as $P_{dynamic}(s)$ of (\ref{eq:tf_ol_low}) in the \textit{dynamic load case} and it can be modeled as $P_{static}(s)$ of (\ref{eq:trq_tf_ol}) in the \textit{static load case}.

It is important to investigate the control input $\tau_c$ and the motor velocity $v_m$ during this force control because it should be checked whether $\tau_c$ and $v_m$ exceed their limiting values or not.

First, the transfer function from the desired torque output $\tau_d$ to the control input $\tau_c$ is derived as (\ref{eq:comp_sens}) based on Fig. \ref{fig:SEA_FCFB}.
\begin{equation}
T_c(s)=\frac{N_m^{-1}C(s)}{1+N_m^{-1}C(s)P(s)}\Tau_d(s)
\label{eq:comp_sens}
\end{equation}
where $T_c(s)$ and $\Tau_d(s)$ are the control input (i.e., motor torque) and the desired torque output in the Laplace domain, respectively. 

Then, the transfer function to the motor velocity $v_m$ under the force control is derived as (\ref{eq:v_m}).
\begin{eqnarray}\label{eq:v_m}
V_m(s)&=&P_V(s)T_c(s)\\ \nonumber
&=&P_V(s)\frac{N_m^{-1}C(s)}{1+N_m^{-1}C(s)P(s)}\Tau_d(s)
\end{eqnarray}
where $V_m(s)$ is the motor velocity in the Laplace domain.

Note that the configuration in Fig. 3 is general without specifying the type of SEA or the type of controller $C(s)$ as long as they are stable. That is, the proposed MTT, which will be defined based on (\ref{eq:comp_sens}) and (\ref{eq:v_m}) can be applied to any types of SEA and controller.

In this paper, a P or PD controller is utilized as $C(s)$, the gains of which are designed to guarantee the stability of the closed loop system. Even though a more complicated controller can be utilized, a P or PD controller is utilized here as it is the most widely employed controller.
     
\subsection{Maximum Torque Transmissibility based on Maximum Continuous Motor Torque}
Based on the configuration of the force controller in Sec. \ref{sec:controller}, the following three conditions are considered in order to define MTT:
\begin{enumerate}
\item The nominal torque $\Tau_{nominal}$ of a motor is adopted as the maximum continuous motor torque $\Tau_{m.c}$ \cite{maxon, lee2016maximum}.
\item The motor velocity is considered to be restricted when it exceeds the maximum permissible velocity $V_p$ of a motor\cite{maxon, lee2016maximum}.
\item The SEA is controlled to provide the desired torque $\tau_d$ by the feedback control of the spring deformation $\theta_d$ as shown in Fig. \ref{fig:SEA_FCFB} \cite{lee2016maximum}.
\end{enumerate}

The maximum torque output $\Tau_{d}^{max}$ of SEA is supposed to be the product of the maximum continuous motor torque $\Tau_{m.c}$ and the gear ratio $N_{m}$, as follows.
\begin{equation}
\Tau_{d}^{max}=N_{m}\Tau_{m.c}
\label{eq:max_SEA_trq}
\end{equation}
Notice that $\Tau_{d}^{max}$ is the maximum torque value that can be expected from SEA when it works ideally as a transmission system.

The control input $\tau_c$, which is also the motor torque required to achieve the maximum SEA output torque $\Tau_{d}^{max}$ can be derived as follows, based on the relationship between $\tau_d$ and $\tau_c$ in (\ref{eq:comp_sens}).
\begin{equation}
T_{c}(s)=\frac{\left[ 1+K_{s}\left(P_{l}(s)+N_m^{-2}P_{m}(s)\right)\right] C(s)N_{m}^{-1}}{1+K_{s}\left[P_{l}(s)+N_m^{-2}P_{m}(s)(1+C(s))\right]}\Tau_{d}^{max}
\label{eq:required_tauC}
\end{equation}
In this equation, the dynamic load model (\ref{eq:tf_ol_low}) is adopted as the plant model $P(s)$ to describe the dynamics of SEA.

If $\tau_c$ in (\ref{eq:required_tauC}) exceeds the maximum continuous motor torque $\Tau_{m.c}$, SEA cannot generate $\Tau_{d}^{max}$, which leads to the definition of MTT; MTT is defined as the ratio of the magnitude of $\tau_c$ in (\ref{eq:required_tauC}) to the maximum continuous motor torque $\Tau_{m.c}$. By replacing $\Tau_d^{max}$ in (\ref{eq:required_tauC}) with $N_m \Tau_{m.c}$ as in (\ref{eq:max_SEA_trq}), the proposed MTT is finally derived as follows.
\begin{eqnarray}
MTT_{\tau} &=& \frac{  1  } { \Tau_{m.c} }  \left|
 T_c(s)  \right| \nonumber \\
&=&\left| \frac{\left[ 1+K_{s}\left(P_{l}(s)+N_m^{-2}P_{m}(s)\right)\right] C(s)}{1+K_{s}\left[P_{l}(s)+N_m^{-2}P_{m}(s)(1+C(s))\right]} \right|
\nonumber \\
\label{eq:MTT_T}
\end{eqnarray}

Notice that $MTT_{\tau}$ in (\ref{eq:MTT_T}) has the following features
\begin{enumerate}
  \item $MTT_{\tau}$ is a non-dimensional functional value of $s$.
  \item $MTT_{\tau}$ is interpreted as the required control input normalized by $\Tau_{m.c}$.
  \item $MTT_{\tau}$ can be utilized to analyze the frequency characteristic of the transmissibility of maximum torque.
\end{enumerate}

The magnitude of 1 dB or 0 dB of $MTT_{\tau}$ represents the critical level; if the magnitude of $MTT_{\tau}$ is larger than 1 dB or 0 dB at a certain frequency, it means that the motor cannot produce the required torque for SEA to generate the maximum torque $\Tau_{d}^{max}$ in that frequency, which may cause force generation error or even instability. 

\subsection{Maximum Torque Transmissibility based on Maximum Permissible Velocity}
As explained in Sec. \ref{sec:intro}, the maximum velocity of the motor is also limited by the drive system, which should be taken into account in MTT \cite{lee2016maximum}. To this end, the motor velocity output to achieve the maximum desired torque output $\Tau_d^{max}$ is calculated using (\ref{eq:v_m}) and (\ref{eq:max_SEA_trq}) as follows.
\begin{equation}
V_m(s)=\frac{N_m^{-1}C(s)P_{m}(s)[1+K_sP_l(s)]s}{1+K_{s}\left[P_{l}(s)+N_m^{-2}P_{m}(s)(1+C(s))\right]}\Tau_{d}^{max},
\label{eq:vel_w_max_trq}
\end{equation}

Another Maximum Torque Transmissibility can be defined by assessing this motor velocity with regard to the maximum permissible velocity $V_{p}$, which is given as $MTT_{V}$ in (\ref{eq:MTTv}).
\begin{eqnarray}
MTT_{V}\!  &\! =\! &\! \frac{1}{V_{p}} \left|    V_{m} (s) \right| \nonumber
\\\!&\!=\!&\! \left|\frac{P_{m}(s)C(s)[1+K_sP_l(s)]s}{1+K_{s}\left[P_{l}(s)+N_m^{-2}P_{m}(s)(1+C(s))\right]}\right|\frac{\Tau_{m.c}}{V_{p}}\nonumber \\
\label{eq:MTTv}
\end{eqnarray}
Notice that (\ref{eq:max_SEA_trq}) is also utilized in this derivation.

As $MTT_{\tau}$ in (\ref{eq:MTT_T}), 0 dB of $MTT_{V}$ is the critical level over which the motor torque will be restricted by the motor drive, and subsequently, the desired maximum torque cannot be provided by the SEA.
It is noticeable that $MTT_V$ includes $\Tau_{m.c}$ and $V_{p}$, which are determined mostly by the intrinsic property of a motor, taking into consideration the thermal, electrical and mechanical properties of the motor.

The defined $MTT_\tau$ in (\ref{eq:MTT_T}) and $MTT_V$ in (\ref{eq:MTTv}) contain the load model $P_l(s)$. As explained in Sec. \ref{sec:dyn_mdl_SEA}, $P_l(s)$ can be modeled in various forms considering the environments in which the SEA contacts. In particular, there are many studies on SEA modeling the contact environment with high impedance described in Sec. \ref{sec:dyn_mdl_SEA} as \textit{static load case}. The MTTs of the SEA in the \textit{static load case} are derived by limiting $J_l$ and $B_l$ to infinity, which are given as:
\begin{equation}
MTT_{\tau}^{sta}=\left| \frac{\left[ 1+N_m^{-2}K_{s}P_{m}(s)\right] C(s)}{1+N_m^{-2}K_{s}P_{m}(s)(1+C(s))} \right|,
\label{eq:MTT_T_high}
\end{equation}
\begin{equation}
MTT_{V}^{sta}=\left|\frac{P_{m}(s)C(s)s}{1+N_m^{-2}K_{s}P_{m}(s)(1+C(s))}\right|\frac{\Tau_{m.c}}{V_{p}}.
\label{eq:MTT_V_high}
\end{equation}

\subsection{Maximum Torque Bandwidth}\label{sec:MTT_band}
Utilizing the proposed MTT, a novel frequency bandwidth, the maximum torque frequency bandwidth can be defined, in which the maximum torque generation of SEA is guaranteed. The maximum torque frequency bandwidth can play a key role in the evaluation of mechanical design and controller design as an analysis tool that can indicate the influence of design parameters on MTT.

\begin{figure}
\begin{center}
\includegraphics[width=1\columnwidth]{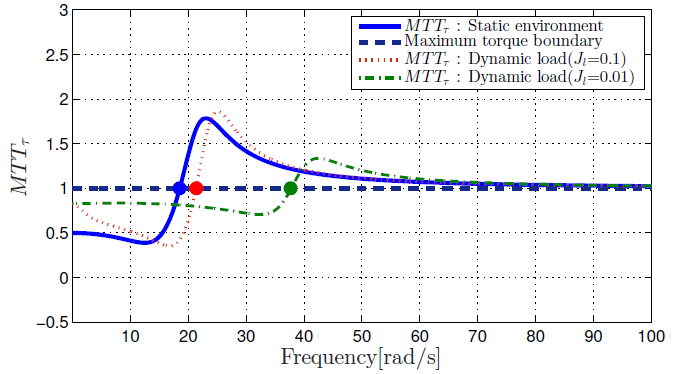}
{\footnotesize (a) $MTT_{\tau}$}
\includegraphics[width=1\columnwidth]{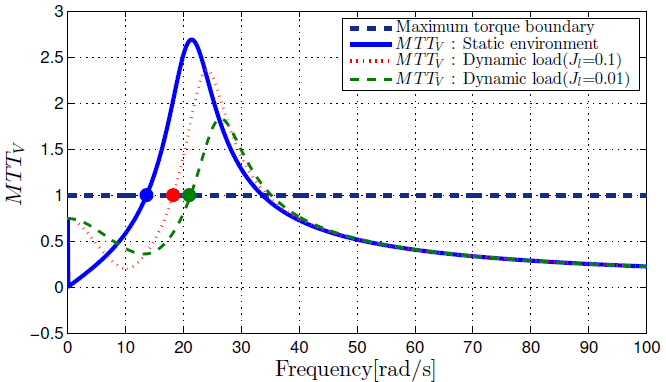}
{\footnotesize (b) $MTT_{V}$}
\caption{\footnotesize (a) $MTT_{\tau}$  and (b) $MTT_V$  with various load inertia conditions including the static case. Other parameters are from Table \ref{tab:SEA_param}, and the feedback controller $C(s)$ is designed as a proportional control with the gain $K_p = 1$.}
\label{fig:MTT_varyingL}
\end{center}
\end{figure}

Figure \ref{fig:MTT_varyingL} (a) shows the $MTT_{\tau}$ with various load conditions.
The maximum torque bandwidth $\omega_{MT_{\tau}}$ in this figure is defined as the lowest frequency where $MTT_{\tau}$ meets 0 dB. This frequency can be calculated using the MTT definition in (\ref{eq:MTT_T}) as
\begin{equation}
\left|\frac{\left[ 1+K_{s}\left(P_{l}(s)+N_m^{-2}P_{m}(s)\right)\right] C(s)}{1+K_{s}\left[P_{l}(s)+N_m^{-2}P_{m}(s)(1+C(s))\right]}\right|_{s=j\omega_{MT_{\tau}}} = 1.
\label{eq:omega_MT_t}
\end{equation}

Figure \ref{fig:MTT_varyingL} (a) shows how the maximum torque bandwidth changes as $J_l$ changes; the critical frequency lowers as $J_l$ increases, and it can be regarded that the static load condition is the poorest case in terms of the maximum torque bandwidth.

$MTT_V$ also defines the frequency bandwidth over which SEA cannot generate the desired maximum torque due to the motor velocity limitation. This bandwidth can be considered the maximum velocity bandwidth $\omega_{MT_V}$, where the motor is required to rotate at its maximum speed to provide $\Tau^d_{max}$. $\omega_{MT_V}$ can be derived using the proposed $MTT_V$ in (\ref{eq:MTTv}) as follows.
\begin{equation}
\left|\frac{C(s)P_{m}(s)[1+K_sP_l(s)]s}{1+K_{s}\left[P_{l}(s)+N_m^{-2}P_{m}(s)(1+C(s))\right]}\frac{\Tau_{m.c}}{V_{p}}\right|_{s=j\omega_{MT_V}} = 1
\label{eq:omega_MT_v}
\end{equation}

Figure \ref{fig:MTT_varyingL} (b) shows $MTT_{v}$ with various load conditions including the static case. In this figure, $\omega_{MT_V}$ is defined as the lowest frequency where $MTT_{v}$ meets 0 dB, which changes with varying $J_l$. Similar to $\omega_{MT_{\tau}}$, $\omega_{MT_V}$ decreases when $J_l$ increases. 

The decrease in the maximum torque frequency bandwidth with regard to the increasing load inertia is due to the decrease in the plant frequency bandwidth given in (\ref{eq:tf_ol_low}), namely, the load condition affects the maximum torque frequency bandwidth mostly in a similar way it affects the plant frequency bandwidth. This is not the case with other parameters, which is elaborated in Sec \ref{sec:frB_exp}.

Between $\omega_{MT_{\tau}}$ and $\omega_{MT_V}$, the smaller frequency is a more critical condition for maximum torque generation of SEA, and thus the final maximum torque frequency bandwidth is determined as (\ref{eq:MTT_decision}).
\begin{equation}
\omega_{MT}=min(\omega_{MT_\tau},\omega_{MT_V})
\label{eq:MTT_decision}
\end{equation}

\section{Verification of MTT through Experiments Using Varying Gear Transmission}\label{sec:exp_MTT}

In this section, the proposed MTT is verified through various experiments, and it is demonstrated that the proposed MTT can be utilized to analyze the force generation performance of SEA.

\subsection{Experimental Setup with Varying Gear Transmission}
\begin{figure}[b!]
\centering
   \includegraphics[width=\columnwidth,]{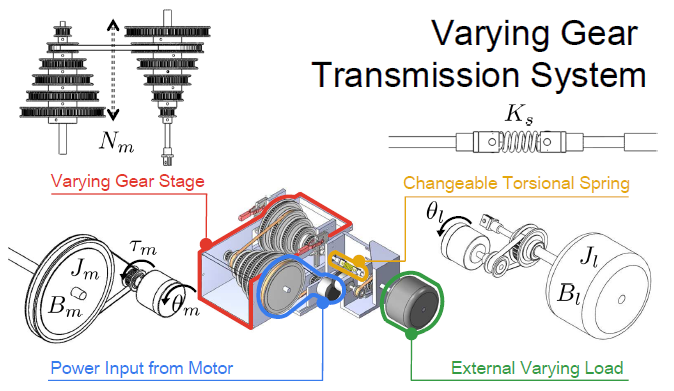}
\caption{Varying Gear Transmission for MTT experiments}
\label{fig:VGT}
\end{figure}
\begin{table}[b!]
\centering
\caption{Identified system parameters.}\label{tab:SEA_param}
\begin{tabular}{c|c|c}
  \hline \hline
 Parameters & Notations & Identified value \\
  \hline \hline
  {Motor inertia} & $J_{m}$ & 0.000075 kg$\cdot$ m$^{2}$  \\
   \hline
  {Load inertia} & $J_{l}$ & 0.005 kg$\cdot$ m$^{2}$\\
   \hline
  {Motor damping} & $B_{m}$ & 0.0006 N$\cdot$ s/m  \\
   \hline
  {Load damping} & $B_{l}$ & 0.08 N$\cdot$ s/m \\
   \hline
  {Spring stiffness} & $K_{s}$ & 1.1 N$\cdot$ m /rad  \\
   \hline
  {Maximum continuous torque} & $\Tau_{m.c}$ & 0.0315 N$\cdot$m   \\
   \hline
  {Maximum permissible velocity} & $V_{p}$ & 10.472 rad/s  \\
   \hline\hline
\end{tabular}
\end{table}

 In order to investigate whether the derived $MTT_{\tau}$ and $MTT_v$ can identify the performance limitation precisely under various conditions of SEA and how MTT reflects the effect of the mechanical parameter of SEA, a varying gear ratio transmission set is developed and utilized in the following case study.

Figure \ref{fig:VGT} is SEA with varying reduction gears, which is called Varying Gear Transmission (VGT). It consists of a motor with a varying gear stage composed of timing pulleys with timing belts (for low backlash and friction) and a spring. The power source of VGT is a Maxon BLDC motor with the fixed 6:1 reduction ratio. An encoder to measure the motor angle $\theta_m$ is attached to the motor.

The second gear stage is the varying gear stage that can shift its ratio from 6:1 to 1:6 by changing the timing belt, which means the whole gear ratio from the motor to the spring changes from 1:1 to 36:1 (1:1, 2.4:1, 4.5:1, 8:1, 15:1, and 36:1) depending on the position of the second timing belt. This varying gear ratio corresponds to $N_m$ in Fig. \ref{fig:SEA_block}. The shaft of the second gear is connected to a spring with stiffness $K_s$ and the external load, which corresponds to $J_l$ and $B_l$, can be attached to the other end of the spring.

In order to obtain the dynamic model (\ref{eq:trq_tf_ol}) of the experimental setup, the frequency response from the motor torque to the spring torque is measured using an FFT analyzer (ONO-SOKKI CF-9400). Table \ref{tab:SEA_param} shows the parameters of the setup estimated in this way.

\subsection{Verification of MTT in the Static Load Case}\label{sec:frB_exp}
Experiments have been conducted to verify the following points with the load side fixed to simulate the static load case.
 \begin{enumerate}
  \item Reliability of the proposed Maximum Torque Transmissibility with various mechanical parameters
  \item The maximum torque bandwidth $\omega_{MT}$ as an analysis tool to evaluate SEA design parameters
\end{enumerate}

To verify the proposed MTTs in (\ref{eq:MTT_T}) and (\ref{eq:MTTv}) which are functions of frequency, SEA with the VGT is controlled to follow desired torques while the load side of the VGT is fixed. Chirp signals were employed as the desired torque reference, and the motor torque $\tau_c$ and the motor velocity were measured and normalized by $\Tau_{m.c}$ and $V_p$, respectively. The magnitude of the chirp signals were determined based on (\ref{eq:max_SEA_trq}), which change according to the selected gear ratio. The frequency range of the chirp signals was set from 0 rad/s to 50 rad/s. A proportional force feedback control with the gain $K_p$ was employed as $C(s)$ in the experiments. The normalized torque and velocity measurements are compared with the proposed $MTT_{\tau}$ and $MTT_V$ in (\ref{eq:MTT_T_high}) and (\ref{eq:MTT_V_high}),respectively, to verify that they precisely portray the transmissibility or the SEA.

\begin{figure}
\begin{center}	
\includegraphics[width=1\columnwidth]{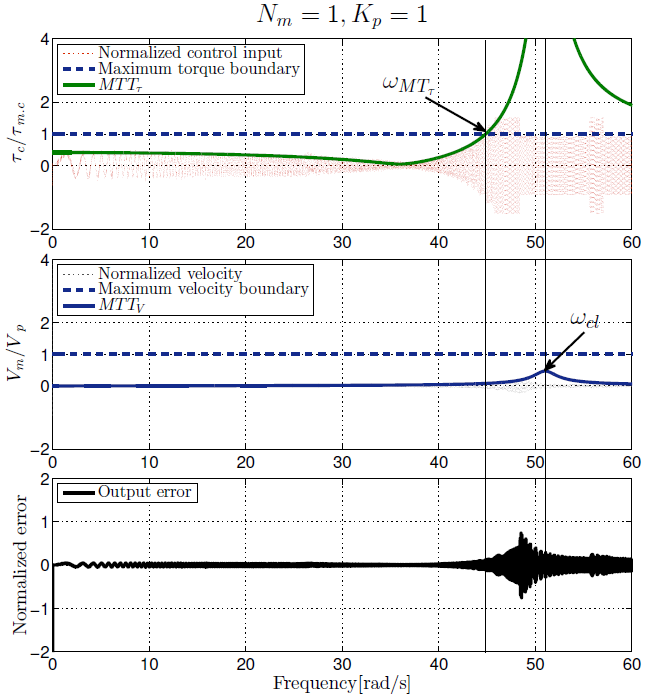}
{\footnotesize (a) Normalized motor input and velocity with $MTT_{\tau}$ and $MTT_V$ ($N_{m}=1$)}
\includegraphics[width=1\columnwidth]{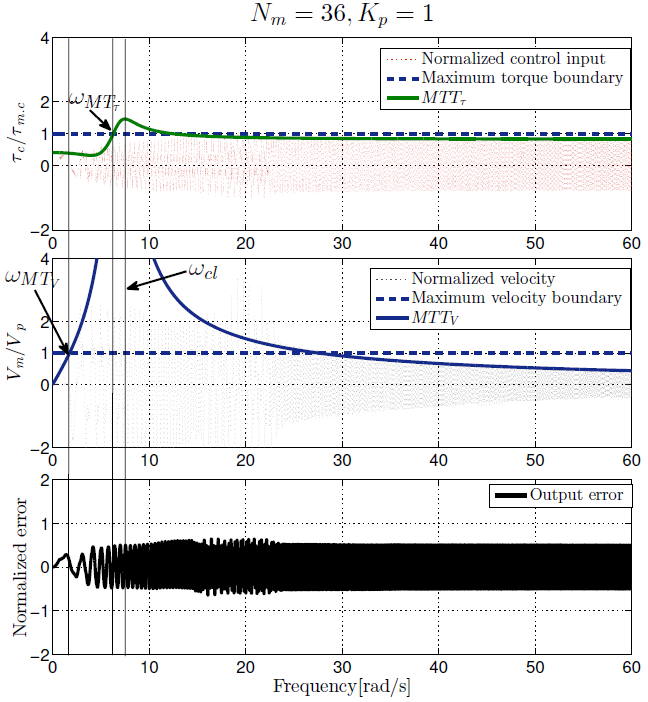}
{\footnotesize (b) Normalized motor input and velocity with $MTT_{\tau}$ and $MTT_V$ ($N_{m}=36$)}
\caption{Experimental results of MTT verification with various gear ratios}
\label{fig:exp_MTT}
\end{center}
\end{figure}

Experiments were conducted with two different sets of reduction gear ratio, which are 1:1 and 36:1. Figures \ref{fig:exp_MTT} (a) and (b) show the experimental results. The thin red dotted lines are the measured control input $\tau_{c}$ normalized by $\Tau_{m.c}$ and the bold solid green lines are $MTT_{\tau}$ calculated from (\ref{eq:MTT_T}) in the upper graphs of (a) and (b). The thin black dotted lines are the velocity output normalized by $V_p$, and the bold solid blue lines are $MTT_{V}$ from (\ref{eq:MTTv}) in the middle graphs. The dashed lines in all the graphs represent the maximum boundary for torque and velocity. Lastly, the black bold lines in the lower graphs are torque control errors normalized by the desired torque output $\Tau_d^{max}$.

From these results, it is verified that the proposed $MTT_{\tau}$ and $MTT_{V}$ can successfully portray the required motor torque and velocity, respectively, and thus can be utilized to express maximum torque transmissibility. Even though the motor driver occasionally allows the motor to generate torque over the maximum continuous torque (based on motor temperature and duration time), the measured actual motor torques were mostly bounded by the maximum continuous motor torque.

Restriction by the maximum permissible velocity is done in a different way from the torque limitation: when the motor velocity exceeds maximum permissible velocity, the velocity is not directly restricted to be the maximum value, but the current or the motor torque decreases instead. This is why the actual normalized motor velocity can occasionally go over 1, the maximum permissible velocity level in Fig. \ref{fig:exp_MTT} (b). However, due to the limitation, the motor torque decreases when the motor velocity becomes greater than the maximum permissible velocity, and thus cannot reach the required level, as can be verified by an examination of the normalized motor input in Fig. \ref{fig:exp_MTT} (b).

Any of both limitations causes large torque control error from the desired torque output: when the gear ratio $N_m$ is small ($N_m = 1$), $\omega_{MT_{\tau}}$ determines the maximum torque bandwidth, beyond which the motor torque is limited and the torque error increases as shown in Fig. \ref{fig:exp_MTT} (a). As the gear ratio becomes large ($N_m = 36$), $\omega_{MT_V}$ becomes smaller than $\omega_{MT_{\tau}}$ and determines the maximum torque bandwidth as shown in Fig. \ref{fig:exp_MTT} (b).

Large torque errors caused by the limitations in this experiment imply that the frequency bandwidth estimated without consideration of the motor limit can be erroneous. On the other hand, the proposed maximum torque bandwidth can precisely indicate the frequency up to which the torque control performance is guaranteed and the maximum torque transmission is achieved.

The MTT for the dynamic load case is verified through the experiments in the following section.

\section{Discussion- Novel Bandwidth Criterion using MTT} \label{sec:MTT_omega}
In this section, it is shown that MTT, in particular the maximum torque bandwidth $\omega_{MT}$, can be utilized as a criterion to evaluate the mechanical and control design parameters of SEA.

In the dynamic case where the torque of SEA is transmitted to the load
and generates dynamic motions, MTT becomes more complicated and it becomes
necessary to take load dynamics into account. Even under the dynamics case,
the proposed large torque bandwidths can be calculated as
(\ref{eq:omega_MT_t}) and (\ref{eq:omega_MT_v}), which are functions of the
mechanical design parameters $ N_m $, $ K_s $, control parameters $ K_p
$, $ K_d $, and load environment parameters $J_l$, $B_l$.

As large torque bandwidth $\omega_{MT}$ is a metric to indicate the
performance of SEA as a transmission, it can offer novel insights on how the
parameters affect the large force transmissibility and be utilized as a guideline for the design of SEA; the following three points are new findings that can be drawn only through the proposed maximum force bandwidth.

\begin{enumerate}
  \item There is a certain force feedback gain $K_p$ that decreases $\omega_{MT}$ abruptly
      (discussed in Sec. \ref{sec:MTT_w_fb_gain})
  \item There is a certain gear ratio $N_m$ from which the velocity limit becomes a more
      significant limitation for maximum force generation (discussed in
      Sec. \ref{sec:MTT_w_gear_ratio})
  \item There is a certain spring constant $K_s$ that decreases $\omega_{MT}$ abruptly (
      discussed in Sec. \ref{sec:MTT_w_stiffness})
\end{enumerate}

The parameters $K_p$, $N_m$, and $K_s$ are usually set as high as possible for better force generation performance. In this session, however, it is shown that they are to be limited in terms for MTT and the bandwidth $\omega_{MT}$. This point is demonstrated through analytical
discussions and experiments.

The parameters utilized in the following subsections are from Table
\ref{tab:SEA_param} with the default gear ratio set to $N_m=8$ and the
default PD controller ($C(s) = K_p + K_ds$) gains are set to $K_p=0.8$ and
$K_d=0.05$. Notice that $K_p$ and $K_d$ used in this section are selected using Matlab graphical tuning method to have a closed loop control bandwidth of 5 Hz. 
\subsection{Limitation of Feedback Gain in Terms of $\omega_{MT}$} \label{sec:MTT_w_fb_gain}
\begin{figure}[t!]
\begin{center}	
\includegraphics[width=.48\columnwidth]{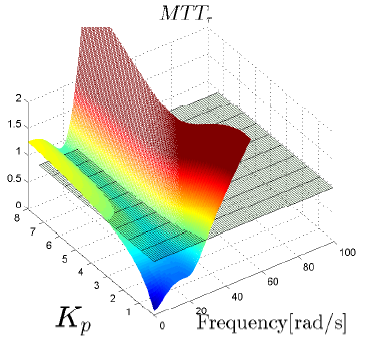}
\includegraphics[width=.48\columnwidth]{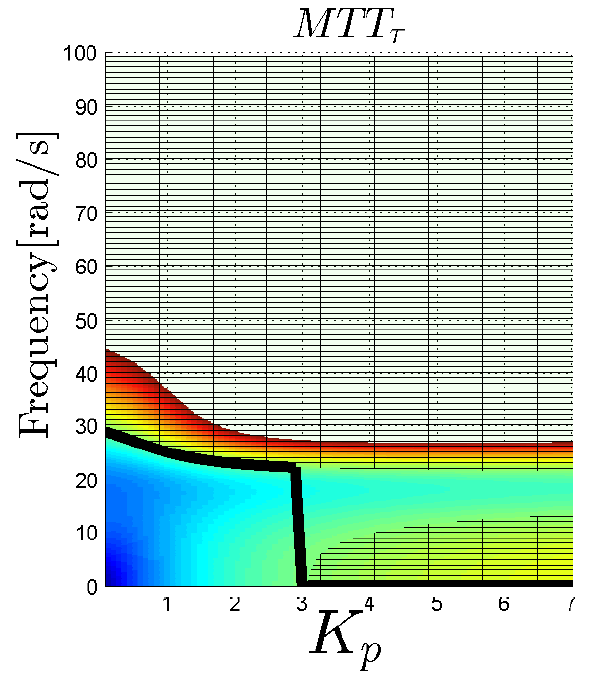}
{\footnotesize(a) $MTT_\tau$ and $K_p$    \quad \quad \quad \quad \quad \quad
\quad (b) $\omega_{MT_{\tau}}$ and $K_p$} 
\caption{\footnotesize $MTT_\tau$
and $\omega_{MT_{\tau}}$ with regard to various $K_p$ values.}
\label{fig:MTT_wrt_gain}
\end{center}
\end{figure}
\begin{figure}[t!]
\begin{center}	
\includegraphics[width=0.9\columnwidth]{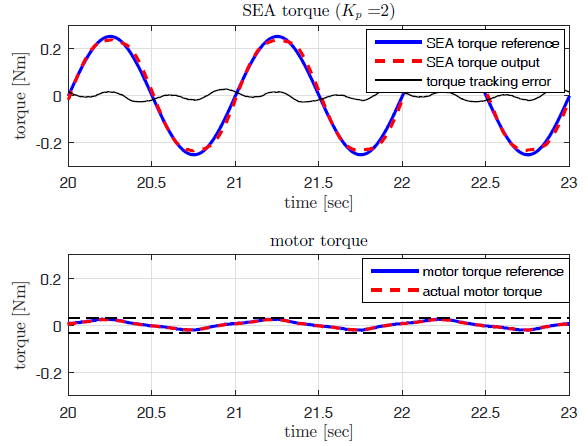} \\
{\footnotesize (a) $K_p=2$}
\includegraphics[width=0.9\columnwidth]{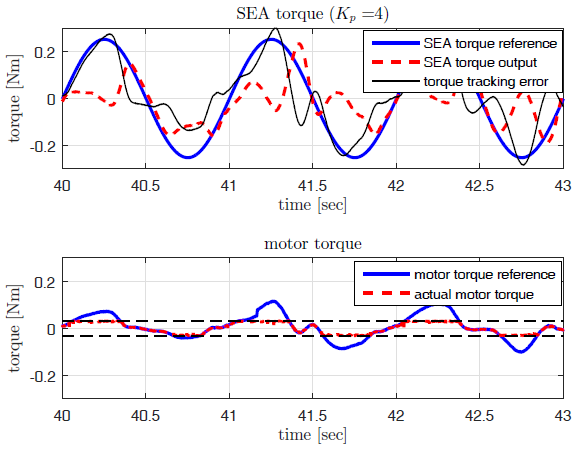} \\
{\footnotesize (b) $K_p=4$}
\caption{\footnotesize Maximum torque tracking performance comparison between two different gains}
\label{fig:exp_t_Kp}
\end{center}
\end{figure}
Figure \ref{fig:MTT_wrt_gain} (a) shows the three dimensional plot of
$MTT_{\tau}$ with respect to the change in gain $K_p$ and frequency
$\omega$. As explained in Sec. \ref{sec:MTT_band}, SEA cannot generate
the maximum torque in the area where $MTT_{\tau}$ exceeds 1 in Fig.
\ref{fig:MTT_wrt_gain}. The bottom view of the three dimensional
$MTT_{\tau}$, given in Fig. \ref{fig:MTT_wrt_gain} (b), clearly displays how the maximum torque bandwidth $\omega_{MT_{\tau}}$ changes with regard to $K_p$.

The area where the grid is disclosed in Fig. \ref{fig:MTT_wrt_gain} (b) is
where SEA cannot generate the maximum torque and $\omega_{MT_{\tau}}$ is determined based on this area. The thick solid line in this figure
indicates the relationship between $\omega_{MT_{\tau}}$ and $K_p$, which
shows that {$\omega_{MT_{\tau}}$} drops to 0 Hz from a certain gain $K_p$.

This relationship implies that the maximum torque generation performance
deteriorates abruptly when the gain $K_p$ is set too high. Figure
\ref{fig:exp_t_Kp} shows experimental results to verify this point; SEA is
controlled to generate the maximum torque at 1 Hz frequency with two different gains. $K_p$ is set to 2 in the left case, and $K_p$ is set to 4 in the right case.

Contrary to common sense, the results show that tracking performance
deteriorates with higher feedback gain.

The proposed equation (\ref{eq:omega_MT_t}) can be utilized to precisely examine the relationship between the gain, and $\omega_{MT_{\tau}}$ and $K_p$. For example, the gain value from which $\omega_{MT}$ suddenly drops to 0 Hz can be specified from the DC component of (\ref{eq:omega_MT_t}) as follows.
\begin{equation}
K_p = 1+N_m^{-2}\frac{B_l}{B_m}
\label{eq:MTT_dc}
\end{equation}
This value can be utilized as marginal gain when the large torque generation
performance is considered significant.

\subsection{Influence of Gear Ratio on Dynamic Maximum Torque Generation} \label{sec:MTT_w_gear_ratio}
\begin{figure}[t!]
\begin{center}	
\includegraphics[width=.48\columnwidth]{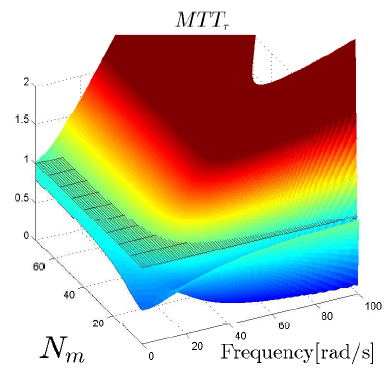}
\includegraphics[width=.48\columnwidth]{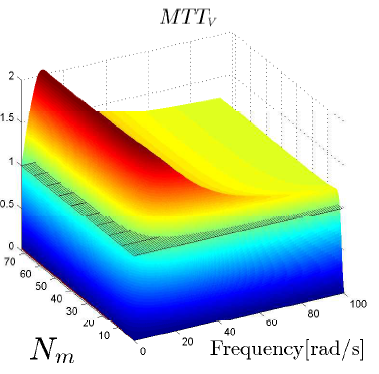}
{\footnotesize(a) $MTT_\tau$  \quad \quad \quad \quad \quad \quad \quad (b)  $MTT_V$ }
\caption{\footnotesize $MTT_\tau$ and $MTT_V$ with regard to $N_m$}
\label{fig:MTT_w_gear}
\end{center}
\end{figure}
\begin{figure}[t!]
\begin{center}	
\includegraphics[width=.8\columnwidth]{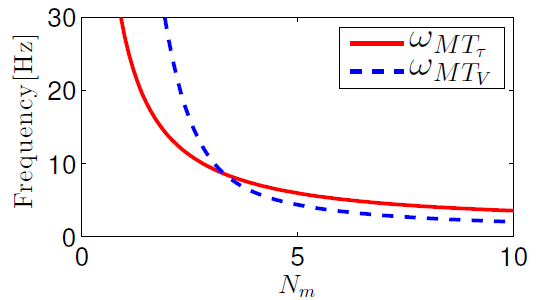}
\caption{\footnotesize Large torque bandwidth $\omega_{MT_\tau}$(solid red) and $\omega_{MT_V}$(dashed blue) with regard to $N_m$} \label{fig:MTT_omega_w_Nm}
\end{center}
\end{figure}

The gear ratio in the SEA is usually set large to generate a large torque.
However, the proposed MTT analysis reveals that large gear ratio reduces the
maximum torque bandwidth $\omega_{MT}$, which means a large gear ratio can
increase the magnitude of the torque output while it decreases the response
time of the large torque.

Figure \ref{fig:MTT_w_gear} shows the three dimensional plots of $MTT_\tau$
and $MTT_V$ with regard to gear ratio $N_m$ and frequency $\omega$,
where it can be found that both bandwidths $\omega_{MT_\tau}$ and
$\omega_{MT_V}$ decrease as $N_m$ increases.

Although it is well understood that the velocity limitation of a motor
becomes a more significant limitation whenever a very large reduction gear is
employed, there has not been a clear standard indicating the gear ratio from
which the velocity limitation plays a significant role. The proposed large
torque bandwidth can be utilized to assess whether the torque limit is
critical or the velocity limit is critical.

Figure \ref{fig:MTT_omega_w_Nm} shows the relationship among
$\omega_{MT_\tau}$, $\omega_{MT_V}$, and $N_m$. The lower graph between the two in Fig. \ref{fig:MTT_omega_w_Nm} is the dominant maximum torque bandwidth as shown in (\ref{eq:MTT_decision}), which can identify from what gear ratio the velocity limitation becomes the more dominant factor. This assessment method can be utilized when selecting a motor and gear ratio in terms of maximum torque generation.

\subsection{Assessment of Spring Stiffness in Terms of $\omega_{MT}$} \label{sec:MTT_w_stiffness}

It is generally accepted knowledge that the natural frequency of SEA
increases as the spring constant becomes large, which can thus enhance the
bandwidth of force control. The maximum torque bandwidth $\omega_{MT}$ reveals a different aspect of large spring constants.

\begin{figure}[t!]
\begin{center}	
\includegraphics[width=.48\columnwidth]{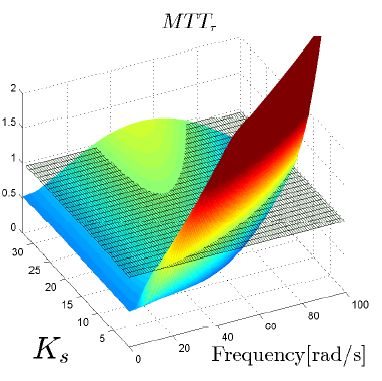}
\includegraphics[width=.48\columnwidth]{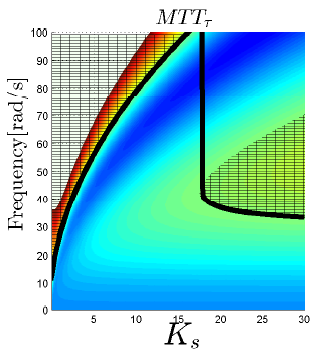}
{\footnotesize(a) $MTT_\tau$ and $K_s$    \quad \quad \quad \quad \quad \quad
\quad (b) $\omega_{MT_{\tau}}$ and $K_s$ } \caption{\footnotesize $MTT_\tau$
and $\omega_{MT_{\tau}}$ with regard to $K_s$} \label{fig:MTT_w_Ks}
\end{center}
\end{figure}

The three dimensional plot of $MTT_{\tau}$ with regard to the spring
stiffness $K_s$ is given in Fig. \ref{fig:MTT_w_Ks} (a), and the bottom view
of the three dimensional plot is Fig. \ref{fig:MTT_w_Ks} (b), which displays
the relationship between the large torque bandwidths $\omega_{MT_{\tau}}$ and
$K_s$.

As earlier explained, SEA cannot provide the maximum torque in the gridded area in Fig. \ref{fig:MTT_w_Ks} (b), and the frequency where the gridded area starts is the bandwidth $\omega_{MT_{\tau}}$. The thick solid line in this figure indicates the relationship between $\omega_{MT_{\tau}}$ and $K_s$, which shows that 1) $\omega_{MT_{\tau}}$ increases as $K_s$ increases, and 2) too large a spring stiffness suddenly drops $\omega_{MT_{\tau}}$. The second point shows that a high spring coefficient does not always increase the
bandwidth, which is contrary to common knowledge.

\begin{figure}[t!]
\begin{center}	
\includegraphics[width=.48\columnwidth]{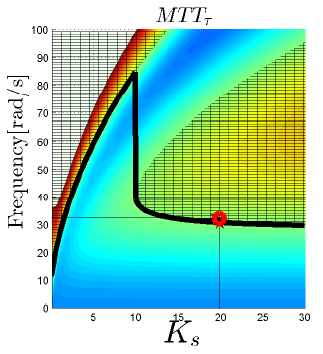}
\includegraphics[width=.48\columnwidth]{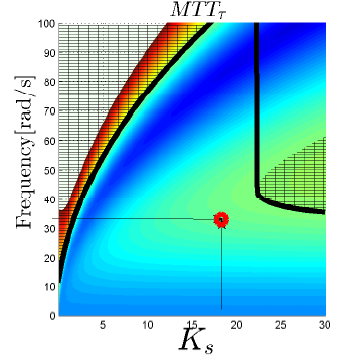}

{\footnotesize (a) Low load inertia }{\tiny $J_l=0.003$ kg$\cdot$m$^{2}$}
{\footnotesize(b) High load inertia }{\tiny$J_l=0.007$ kg$\cdot$m$^{2}$}

\caption{\footnotesize  $\omega_{MT_{\tau}}$ with regard to $K_s$ under two
different load inertia conditions} \label{fig:MTT_w_Ks_load}
\end{center}
\end{figure}

The spring coefficient where $\omega_{MT_{\tau}}$ suddenly drops varies
depending on the load condition. Figure \ref{fig:MTT_w_Ks_load} illustrates the relationships between $\omega_{MT_{\tau}}$ and $K_s$ under two load conditions: the left plot is with low load inertia ($J_l = 0.003$ kg$\cdot$m$^{2}$), and the right plot is with high load inertia ($J_l = 0.007$ kg$\cdot$m$^{2}$).
From the comparison between these two plots, it can be concluded that large
load inertia can increase large torque bandwidth whenever the same spring
stiffness is employed.

To validate this aspect experimentally, maximum torque tracking control is
applied to the VGT under two different load conditions: $J_l = 0.003$
kg$\cdot$m$^{2}$ and $J_l = 0.007$ kg$\cdot$m$^{2}$. The spring of 20 N·m/rad
is utilized for this experiment, and other parameters are set the same as in
Table \ref{tab:SEA_param}. The reference torque is set to a sinusoidal signal
with the frequency of 5 Hz and the maximum magnitude of 0.252 N·m
($=\Tau_{m.c}\cdot N_m$)
\begin{figure}
\begin{center}
\includegraphics[width=0.9\columnwidth]{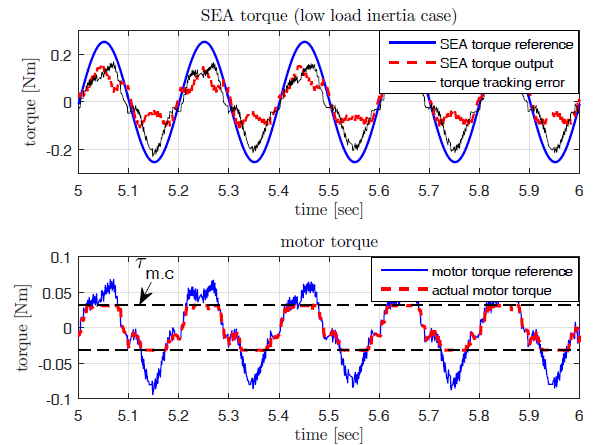}\\
{\footnotesize {(a) Low load inertia }}{\tiny $J_l=0.003$ kg$\cdot$m$^{2}$}\\
\includegraphics[width=0.9\columnwidth]{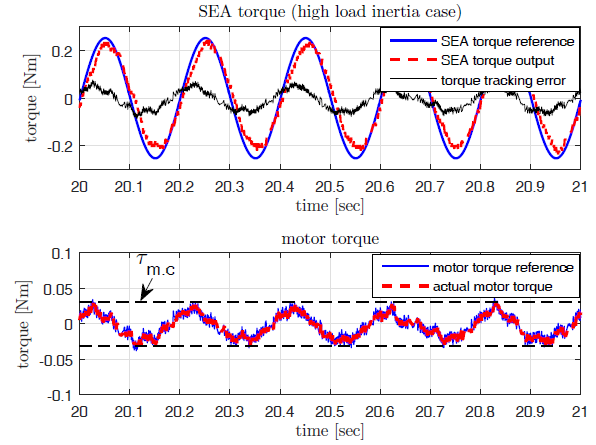}\\
{\footnotesize {(b) High load inertia} }{\tiny$J_l=0.007$ kg$\cdot$m$^{2}$}\\
\caption{\footnotesize  Large torque tracking performance with two load
conditions.} \label{fig:exp_w_Ks_load}
\end{center}
\end{figure}

The stiffness in this experiment $K_s=20$N·m/rad corresponds to the red thickly drawn circles in Fig. \ref{fig:MTT_w_Ks_load}. Even though the stiffness is set the same in both cases, it is in the gridded area in Fig. \ref{fig:MTT_w_Ks_load} (a), and the experimental result in Fig. \ref{fig:exp_w_Ks_load} (a) validates that SEA cannot generate the maximum torque with low load inertia. On the other hand, the same stiffness is
outside the gridded area in Fig. \ref{fig:MTT_w_Ks_load} (b), and
correspondingly, SEA can generate the maximum torque in Fig.
\ref{fig:exp_w_Ks_load} (b).

It is verified through the experiments that the proposed MTT and
$\omega_{MT}$ can precisely represent the maximum torque transmissibility as
a dynamic characteristic of SEA, and thus can be utilized as a guideline
and standard for various mechanical/controller parameters when designing SEA.

\section{Conclusion}
\label{sec:conclusion}
In this paper, Maximum Torque Transmissibility (MTT) is proposed as a mathematical tool to assess the ability of SEA to fully utilize the maximum continuous motor torque. Moreover, the influence of load condition, mechanical parameters, and controller gains on the $\omega_{MT}$ is analyzed using the proposed MTT.

The discussion of the experiments given in this paper can be summarized as follows.
\begin{enumerate}
\item Maximum Torque Transmissibility ($MTT_\tau$ and $MTT_v$) is proposed as a mathematical tool to analyze the influence of the maximum continuous motor torque and the maximum permissible velocity of a motor on SEA performance.
\item Effectiveness of the proposed MTT is verified through experiments under various conditions.
\item Maximum torque frequency bandwidth $\omega_{MT}$ can be derived based on the proposed MTT, in which the transmission of the maximum motor torque by the SEA is guaranteed.
\item It is shown that the proposed $\omega_{MT}$ can be a novel criterion to evaluate SEA design factors such as control gain, gear ratio, and spring stiffness.
\item Experimental results verify the above.
\end{enumerate}

\bibliography{reference}{}
\bibliographystyle{ieeetr}

\end{document}